\documentclass[nohyperref]{article}

\usepackage{microtype}
\usepackage{graphicx}
\usepackage{booktabs} 

\usepackage{hyperref}

 \usepackage[accepted]{wfvml2022}

\usepackage{amsmath}
\usepackage{amssymb}
\usepackage{mathtools}
\usepackage{amsthm}

\usepackage[capitalize,noabbrev]{cleveref}

\theoremstyle{plain}

\theoremstyle{definition}

\theoremstyle{remark}

\usepackage[textsize=tiny]{todonotes}

\usepackage[acronym,nowarn,section,nogroupskip,nonumberlist,nohypertypes={acronym,notation}]{glossaries}
\newacronym{ibp}{IBP}{Interval Bound Propagation}
\newacronym{colt}{COLT}{Convex Layer-wise Adversarial Training}

\usepackage{accents}
\newcommand{\ubar}[1]{\underaccent{\bar}{#1}}

\usepackage{amsmath}
\usepackage{amsfonts}
\usepackage{dsfont}
\usepackage{bm}
\usepackage{xcolor}
\usepackage{etoolbox}
\usepackage{nicefrac}

\newcommand{\mbf}[1]{\mathbf{#1}}

\newcommand{\yb}{\mbf{y}}

\newcommand{\mub}{\bm{\mu}}

\newcommand{\lambdab}{\bm{\lambda}}

\newcommand{\bb}{\mbf{b}}
\newcommand{\ub}{\mbf{u}}
\newcommand{\lb}{\mbf{l}}
\newcommand{\ubhat}{\hat{\mbf{u}}}
\newcommand{\lbhat}{\hat{\mbf{l}}}
\newcommand{\xb}{\mbf{x}}

\newcommand{\balpha}{\bm{\alpha}}
\newcommand{\bbeta}{\bm{\beta}}

\newcommand{\thetab}{\bm{\theta}}

\newcommand{\xbhat}{\mbf{\hat{x}}}

\DeclareMathOperator*{\EX}{\mathbb{E}}

\usepackage{enumitem}

\usepackage{stmaryrd}

\usepackage[labelformat=simple]{subcaption}

\usepackage{booktabs}
\usepackage{adjustbox}
\usepackage{multirow}
\newcommand\Tstrut{\rule{0pt}{2.6ex}}       
\newcommand\Bstrut{\rule[-0.9ex]{0pt}{0pt}} 
\newcommand{\TBstrut}{\Tstrut\Bstrut} 
\renewcommand{\bfseries}{\fontseries{b}\selectfont}
\newrobustcmd{\B}{\bfseries}
\usepackage[normalem]{ulem}
\robustify\uline
\def\Decimal{.000}

\def\Ulinehelp#1.#2 {%
	#1.#2\setbox0=\hbox{#1\Decimal}\hspace{-\wd0}{\if\relax#2\relax%
		\uline{\phantom{#1.0}}\else\uline{\phantom{#1.#2}}\fi}%
}
\usepackage{siunitx}

\wfvmltitlerunning{IBP Regularization for Verified Adversarial Robustness}

\begin{document}

\twocolumn[
\wfvmltitle{IBP Regularization for Verified Adversarial Robustness via Branch-and-Bound}




\begin{wfvmlauthorlist}
\wfvmlauthor{Alessandro De Palma}{oxf}
\wfvmlauthor{Rudy Bunel}{dm}
\wfvmlauthor{Krishnamurthy (Dj) Dvijotham}{gb}
\wfvmlauthor{M. Pawan Kumar}{dm}
\wfvmlauthor{Robert Stanforth}{dm}
\end{wfvmlauthorlist}

\wfvmlaffiliation{oxf}{University of Oxford}
\wfvmlaffiliation{dm}{DeepMind}
\wfvmlaffiliation{gb}{Google Brain}

\wfvmlcorrespondingauthor{Alessandro De Palma}{adepalma@robots.ox.ac.uk}

\wfvmlkeywords{Machine Learning, Formal Verification}

\vskip 0.3in
]



\printAffiliationsAndNotice{}  

\begin{abstract}
	Recent works have tried to increase the verifiability of adversarially trained networks by running the attacks over domains larger than the original perturbations and adding various regularization terms to the objective.
	However, these algorithms either underperform or require complex and expensive stage-wise training procedures, hindering their practical applicability.
	We present IBP-R, a novel verified training algorithm that is both simple and effective. IBP-R induces network verifiability by coupling adversarial attacks on enlarged domains with a regularization term, based on inexpensive interval bound propagation, that minimizes the gap between the non-convex verification problem and its approximations.
	By leveraging recent branch-and-bound frameworks, we show that IBP-R obtains state-of-the-art verified robustness-accuracy trade-offs for small perturbations on CIFAR-10 while training significantly faster than relevant previous work.
	Additionally, we present UPB, a novel branching strategy that, relying on a simple heuristic based on $\beta$-CROWN, reduces the cost of state-of-the-art branching algorithms while yielding splits of comparable quality.
\end{abstract}

\section{Introduction}

The existence of adversarial examples~\citep{Szegedy2014,Goodfellow2015} has raised widespread concerns on the robustness of neural networks. 
As a consequence, many authors have promptly devised algorithms to formally prove the robustness of trained networks~\citep{Katz2017,Ehlers2017,Bunel2018,Zhang2018,Raghunathan2018}.
At the same time, a number of works have focused on training networks for adversarial robustness: first by defending against specific attacks (adversarial training)~\citep{Madry2018}, then providing formal guarantees about attack-independent robustness (verified training)~\citep{Dvijotham2018a,Wong2018,Mirman2018}. 
The vast majority of verified training methods operate by backpropagating over over-approximations of the network's loss under adversarial perturbations, and obtain state-of-the-art results for large perturbations~\citep{Zhang2020,Xu2020,Lyu2021}.
However, these training schemes are typically unable to benefit from tight over-approximations at verification time, hence requiring relatively large networks to perform at their best.
A recent line of work has better leveraged network capacity by enhancing the verifiability of adversarially-trained networks. These algorithms can exploit tighter over-approximations but they either underperform~\citep{Xiao2019} or require expensive procedures in order to reach state-of-the-art performance on small perturbations~\citep{Balunovic2020}.

\looseness=-1
The recent VNN-COMP-21, an international competition on neural network verification~\citep{vnn-comp-2021} highlighted significant scaling improvements in exact verification algorithms~\citep{HenriksenLomuscio21,gpupoly,improvedbab,betacrown}.
We aim to leverage these developments by presenting IBP-R, a novel and inexpensive verified training algorithm that induces network verifiability by:
(i) running adversarial attacks over domains that are significantly larger than the target perturbations,
(ii) exploiting \gls{ibp}~\citep{Mirman2018,Gowal2018b} to minimize the area of the convex hull of the activations, a commonly employed relaxation within recent verification frameworks.
We show that, in spite of its speed and conceptual simplicity, IBP-R yields state-of-the-art results under small perturbations on CIFAR-10. In particular, under $\ell_\infty$ perturbations of radius $\epsilon_{\text{ver}}=2/255$, networks trained via IBP-R attain, on average:
a verified accuracy of 61.97, a robust accuracy of 66.39 under MI-FGSM attacks~\citep{dong2018boosting} and a natural accuracy of 78.19. In this setting, IBP-R trains in less than a third of the runtime of COLT~\citep{Balunovic2020}.
Furthermore, for $\epsilon_{\text{ver}}=8/255$, IBP-R performs competitively with COLT while almost halving its runtime. 

\looseness=-1
Finally, motivated by the task of evaluating the verifiability of networks trained via IBP-R,
we present a simple and novel branching strategy, named UPB, for complete verification via branch-and-bound~\citep{Bunel2018}. 
UPB leverages dual information from the recent $\beta$-CROWN algorithm~\citep{betacrown} to heuristically rank the quality of the possible branching decisions.
We show that, at a cost equivalent to a single gradient backpropagation through the network, UPB obtains a verification performance comparable to the more expensive and state-of-the-art FSB strategy~\citep{improvedbab}. 

\section{Background} \label{sec:background}

In the following, we will use boldface letters to denote vectors (for example, $\xb$), uppercase letters to denote matrices (for example, $W_k$), and brackets for intervals (for example, $[\lb_k,\ub_k]$). 
Furthermore, we will write $\odot$ for the Hadamard product, $\mathds{1}_{{\mbf{c}}}$ for the indicator vector on condition $\mbf{c}$, $\llbracket \cdot \rrbracket$ for integer ranges, and employ the following shorthand: $[\xb]_+ := \max(\xb, \mbf{0})$.

Let us define the data distribution $\mathcal{D}$, yielding points $(\xb, \yb) \in \mathbb{R}^{d} \times \mathbb{R}^{o}$ and let us denote by $\thetab \in \mathbb{R}^{S}$ the network parameters.
Robust training is concerned with training a neural network $f: \mathbb{R}^S \times \mathbb{R}^{d} \rightarrow \mathbb{R}^o$ so that a given property $P: \mathbb{R}^o \times \mathbb{R}^{o} \rightarrow \{0, 1\}$ is satisfied in a region around each input point $\xb$, denoted $\mathcal{C}(\xb)$. In other words, the parameters~$\thetab$ must satisfy the following:
\begin{equation}
	\xb_{0} \in \mathcal{C}(\xb)\ \implies P(f(\thetab, \xb_{0}), \yb) \quad \forall \enskip (\xb, \yb) \in \mathcal{D}.
	\label{eq:verification}
\end{equation}
\looseness=-1 
In this work, we will focus on robustness to adversarial perturbations around the input images. Specifically, $\mathcal{C}(\xb) := \{ \xb_0 : \left\lVert \xb_0 - \xb  \right\rVert_p \leq \epsilon_{\text{ver}}  \}$ and $P$ amounts to checking that the predicted and ground truth classification labels~match.

\subsection{Neural Network Verification} \label{sec:background-verification}

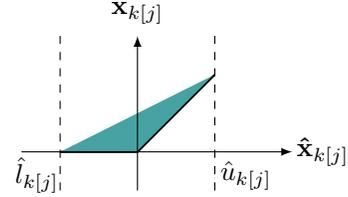
\begin{figure}[t!]
    \vspace{-3pt}
	\centering
	\noindent\resizebox{.28\textwidth}{!}{
		\definecolor{lightgray}{rgb}{0.8,0.8,0.8}

\begin{tikzpicture}
  \tikzset{dummy/.style= {inner sep=0, outer sep=0}}
  \tikzset{cross/.style={circle, draw,
      minimum size=5*(#1-\pgflinewidth),
      inner sep=0pt, outer sep=0pt,
      ultra thick, color=red}}

  \draw[-, thick](-1, 0) to (0, 0) to (1, 1);

  \draw[dashed](-1, -0.5) to (-1, 1.5);
  \draw[dashed](1, -0.5) to (1, 1.5);

  \draw[fill=teal!80, fill opacity=0.9](-1, 0) -- (0,0) -- (1, 1);

  \node[dummy](lb-lab) at (-1.3, -0.3) {$\hat{l}_{k[j]}$\hspace{3pt}};
  \node[dummy](ub-lab) at (1.35, -0.3) {\hspace{3pt}$\hat{u}_{k[j]}$};


  \draw[-latex](-1.5,0) to (2, 0);
  \node[dummy](x-label) at (2.3, 0) {\hspace{3pt} $\xbhat_{k[j]}$};
  \draw[-latex](0,-0.5) to (0, 1.5);
  \node[dummy](x-label) at (0, 1.8) {$\xb_{k[j]}$};

\end{tikzpicture}
	}
	\vspace{-8pt}
	\caption{Convex hull for an ambiguous ReLU~\citep{Ehlers2017}.}
	\vspace{-3pt}
	\label{fig:relucvx}
\end{figure}

Before delving into the task of training a robust network, we first consider the problem of determining whether a given network is robust or not. This involves the formal verification of condition \eqref{eq:verification} on the given network, which is generally NP-HARD~\citep{Katz2017}. Therefore, its exact verification is often replaced by less expensive approximations, which nevertheless provide formal guarantees for a subset of the properties (incomplete verification). 
By means of simple transformations, one can represent both $f$ and $P$ from condition \eqref{eq:verification} via a single network $f'$ of depth $n$~\citep{Bunel2020}, so that incomplete verification corresponds to the following optimization problem:
\begin{equation}
\begin{aligned}
&\smash{\min_{\xb, \xbhat}}\quad \hat{x}_{n} \qquad \text{s.t. } \\[7pt] 
& \xb_0 \in \mathcal{C}(\xb), \\
& \xbhat_{k+1} = W_{k+1} \xb_{k} + \mathbf{b}_{k+1} && k \in \left\llbracket0,n-1\right\rrbracket,\\
& (\xb_k, \xbhat_{k}) \in \text{Rel}(\sigma, \hat{\lb}_k, \hat{\ub}_k) && k \in \left\llbracket1,n-1\right\rrbracket, \\
& \xbhat_k \in [\hat{\lb}_k, \hat{\ub}_k]   && k \in \left\llbracket1,n-1\right\rrbracket,
\end{aligned}
\label{eq:relaxation}
\end{equation}
\looseness=-1
where $\sigma$ denotes the activation function, $W_k$ the weight matrix of the $k$-th layer of $f'$, $\bb_k$ its bias.
The use of $\text{Rel}(\sigma, \hat{\lb}_k, \hat{\ub}_k)$, a convex relaxation of $\sigma$, ensures that problem \eqref{eq:relaxation} is convex, greatly simplifying its solution.
In general, $\text{Rel}(\sigma, \hat{\lb}_k, \hat{\ub}_k)$ is a function of intermediate bounds~$\hat{\lb}_k, \hat{\ub}_k$, which provide ranges on the network pre-activation variables (for details, see appendix \ref{sec:ibs}).
In the context of ReLU networks, which are the focus of this work, a popular relaxation choice is the convex hull of the activation, commonly referred to as the Planet relaxation~\citep{Ehlers2017}.
If $\hat{\lb}_k< 0$ and $\hat{\ub}_k > 0$ (ambiguous ReLU), its shape is given by Figure~\ref{fig:relucvx}. If either $\hat{\lb}_k > 0$ or $\hat{\ub}_k < 0$, the activation is said to be stable and its convex hull can be represented by a line, greatly improving the tightness of the overall network approximation.

When verifying all properties is a requirement (complete verification), problem \eqref{eq:relaxation} is employed as a sub-routine for a global optimization algorithm equivalent to branch-and bound~\citep{Bunel2018}. 
The goal is to find the sign of the minimum of a non-convex problem (reported in appendix~\ref{sec:complete-verification}) whose domain is a subset of the feasible region from problem \eqref{eq:relaxation}.
Complete verifiers hence proceed by recursively splitting the domain (branching) and solving the resulting convex sub-problems (bounding) until a definite answer can be provided. For ReLU activations, the branching is usually performed by splitting the domain of an ambiguous ReLU into its two stable subdomains.

\subsection{Training via the Robust Loss} \label{sec:robust-loss}
In order to enforce condition \eqref{eq:verification} during training, one typically defines a surrogate loss $\mathcal{L} : \mathbb{R}^{o} \times \mathbb{R}^{o} \rightarrow \mathbb{R}$, and seeks to minimize the worst-case empirical risk within~$\mathcal{C}(\xb)$, referred to as the robust loss:
\begin{equation}
    \min_{\thetab}\ \EX_{(\xb, \yb) \in \mathcal{D}} \left[ \max_{\xb' \in \mathcal{C}(\xb)} \mathcal{L} (f(\thetab, \xb'), \yb)\right].
    \label{eq:robust-problem}
\end{equation}
\looseness=-1
The exact computation of $\max_{\xb' \in \mathcal{C}(\xb)} \mathcal{L} (f(\thetab, \xb'), \yb)$ entails the use of a complete verification algorithm ($\S$\ref{sec:background-verification}), which is too expensive to be employed during training. Therefore, the robust loss is typically replaced by an approximation:
adversarial training algorithms~\citep{Madry2018} rely on lower bounds, while certified training algorithms~\citep{Gowal2018b,Zhang2020} employ upper bounds.
Lower bounds are computed by using so-called adversarial attacks: algorithms, such as  PGD~\citep{Madry2018}, that heuristically search for misclassified (adversarial) examples in the input space. Upper bounds are instead computed by solving an instance of problem \eqref{eq:relaxation} where $\text{Rel}(\sigma, \hat{\lb}_k, \hat{\ub}_k)$ typically represents the \gls{ibp} hyper-rectangle~\citep{Mirman2018} or linear bounds on the activation~\citep{Wong2018,Zhang2018}. 
Adversarial training yield models with strong standard accuracy and empirical robustness. However, such robustness is often hard to demonstrate via a formal verification method, and might potentially break under stronger attacks. Stronger robustness guarantees are instead provided by certified training algorithms, at the expense of the standard network accuracy and with longer training times. 

\subsection{Hybrid Training Methods} \label{sec:hybrid}
A line of recent work seeks to bridge the gap between adversarial and certified training by modifying the regions over which the attacks are performed and adding specialized regularization terms.

\citet{Xiao2019} demonstrate that the verified robust accuracy of PGD-trained networks~\citep{Madry2018} can be increased by adding $\ell_1$ regularization and a term encouraging ReLU stability to problem \eqref{eq:robust-problem}:
\begin{equation}
    \min_{\thetab}\ \hspace{-5pt} \EX_{(\xb, \yb) \in \mathcal{D}} \hspace{-2pt} \left[\hspace{-3pt}\begin{array}{l} \max_{\xb' \in \mathcal{C}(\xb)} \mathcal{L} (f(\thetab, \xb'), \yb)\ + \lambda \left\lVert \thetab \right\rVert_1 \\
    + \rho \sum_{j=1}^{k} \tanh(1 - \hat{\lb}_k(\thetab) \odot \hat{\ub}_k(\thetab))^T \mbf{1},  \end{array}\hspace{-8pt}\right.
    \label{eq:relu-stability}
\end{equation}
where $\hat{\lb}_k(\thetab)$ and $\hat{\ub}_k(\thetab)$, which depend on the network parameters, are computed via a tightened version of \gls{ibp}.

\citet{Balunovic2020} propose to employ adversarial training layer-wise, running the attacks over convex outer-approximations of frozen subsets of the network. 
Let us denote by $f^j$ a subset of network $f$ that starts at the $j$-th layer, and by $\thetab^j$ its parameters. 
Furthermore, let $\mathbb{C}_j(\xb)$ represent an outer-approximation of the $j$-th latent space obtained via the zonotope relaxation~\citep{Zhang2018}, relying on zonotope intermediate bounds approximated via Cauchy random projections~\citep{Li2007}.
\gls{colt}, operates on the following objective at the $j$-th stage of the training:
\begin{gather}
	\min_{\thetab^{j}} \hspace{-5pt} \EX_{(\xb, \yb) \in \mathcal{D}} 
	\left\{ \hspace{-5pt} \begin{array}{l}
	 \kappa \left[\begin{array}{l} \max_{\xb' \in \mathbb{C}_j(\xb)} \mathcal{L} (f^{j}(\thetab^{j}, \xb'), \yb) \\ 
	 + \rho_j \left[-\lbhat_{j+1}(\thetab^j)\right]_+^T\left[\ubhat_{j+1}(\thetab^j)\right]_+ \end{array}\right] \\[15pt]
	 + (1 - \kappa) \max\limits_{\xb' \in \mathbb{C}_{j-1}(\xb)} \mathcal{L} (f^{j-1}(\thetab^{j-1}, \xb'), \yb)\\ 
	 + \lambda \left\lVert \thetab^j \right\rVert_1, \end{array}\right.
	 \label{eq:colt}
	 \raisetag{55pt}
\end{gather}
where $[-\lbhat_{j+1}]_+^T\left[\ubhat_{j+1}\right]_+$ is a regularizer for the $(j+1)$-th latent space, inducing ReLU stability and minimizing the area of the zonotope relaxation for ambiguous ReLUs.
As when computing intermediate bounds for $\mathbb{C}_j(\xb)$, the regularizer is computed via approximate zonotope bounds.
In order to gradually transition from one training stage to the other, $\kappa$ is linearly increased from $0$ to $1$ in the first phase of the training. At the first stage ($j=0$), the loss transitions from the natural loss (without any adversarial component) to the regularized PGD loss.
\gls{colt} performs particularly well for smaller perturbation radii, for which it yields state-of-the-art results. However, its complexity and stage-wise nature make it relatively hard to deploy in practice. For instance, \citet{Balunovic2020} employ a different value for both $\rho_j$ and the train-time perturbation radius, which affects both $\mathbb{C}_j(\xb)$ and intermediate bounds, at each training stage.
Further details are provided in appendix~\ref{sec:colt-comparison}.

\section{Training via IBP Regularization} \label{sec:ibpr}

Certified training algorithms that directly employ upper bounds to the robust loss 
\eqref{eq:robust-problem} do not typically benefit from the use of more accurate verification algorithms than those they were trained with~\cite{Zhang2020}. On the other hand, the hybrid training methods from $\S$\ref{sec:hybrid} are designed to be verified with tighter bounds, potentially encoding (part of) the network as a MILP~\citep{Tjeng2019}.
In light of the recent scaling improvements of complete verifiers~\citep{vnn-comp-2021}, we present a robust training method designed for recent branch-and-bound frameworks~\citep{improvedbab,betacrown}, capable of preserving \gls{colt}'s effectiveness while simplifying its training procedure. 

\paragraph{Training objective}
Intuitively, verification is easier if the network is robust by a large margin, and if the employed relaxation accurately represents the network, 
Therefore, we propose a simple training scheme revolving around the following two features: 
(i) adversarial attacks run over significantly larger domains than those employed at test time, 
(ii) a term minimizing the gap between the relaxations used for verification and the original neural network domain.
Given its flexibility, small cost, and widespread usage, we aim to exclusively rely on \gls{ibp} for bounding computations. 
Details for \gls{ibp} can be found in appendix~\ref{sec:ibp}.
We name the resulting method IBP Regularization (IBP-R), which operates on the following objective:
\begin{equation}
	\min_{\thetab} \hspace{-2pt} \EX_{(\xb, \yb) \in \mathcal{D}} 
	 \hspace{-2pt} \left\{ \hspace{-5pt} \begin{array}{l}
		\kappa \left[\begin{array}{l}\max_{\xb \in \mathcal{C}_+(\xb)} \mathcal{L} (f(\thetab, \xb), \yb)\ + \\ \frac{\rho}{2} \sum_{j=1}^{n} \left[-\lbhat_{j}(\thetab)\right]_+^T\left[\ubhat_{j}(\thetab)\right]_+ \end{array}  \right] \\[15pt]
		+\ (1 - \kappa) \mathcal{L} (f(\thetab, \xb), \yb) + \lambda \left\lVert \thetab \right\rVert_1, \end{array}\right.
\label{eq:ibpr}
\end{equation}
where $\frac{1}{2} [-\lbhat_{j}(\thetab)]_+^T\left[\ubhat_{j}(\thetab)\right]_+$ represents the area of the widely-employed ReLU convex hull represented in Figure~\ref{fig:relucvx}, and $\mathcal{C}_+(\xb)$ is a superset of the original input specification from condition \eqref{eq:verification}. 
For adversarial robustness specifications, $\mathcal{C}_+(\xb) := \{ \xb_0 | \left\lVert \xb_0 - \xb  \right\rVert_p \leq \alpha \epsilon_{\text{ver}}  \}$, with $\alpha \geq 1.6$ in our experiments (see $\S$\ref{sec:experiments}). 
Note that we regularize over all the activations of the network at once, with the same regularization coefficient.
An in-depth comparison with COLT is available in appendix~\ref{sec:colt-comparison}.

\paragraph{Regularization masking}
\looseness=-1
When a property holds, the tightness of the employed relaxations is fundamental in order to swiftly provide a formal guarantee via branch-and-bounds methods. On the other hand, verifying that the property does not hold typically implies finding counter-examples via adversarial attacks~\citep{vnn-comp-2021}.
As a consequence, there is no need to encourage tightness by minimizing $\frac{1}{2} [-\lbhat_{j}(\thetab)]_+^T\left[\ubhat_{j}(\thetab)\right]_+$ when $P(f(\thetab, \xb_{0}), \yb)\ \forall\ \xb_{0} \in \mathcal{C}(\xb)$ is unlikely to hold after training.
In order to take this observation into account, we propose to mask the convex hull regularizer when a counter-example is found for the current sample.
Denoting by $\tilde{\xb}$ the point found by the train-time adversarial attack, we perform the following substitution in objective~\eqref{eq:ibpr}:
\begin{equation*}
    \frac{1}{2} [-\lbhat_{j}(\thetab)]_+^T\left[\ubhat_{j}(\thetab)\right]_+ \hspace{-2pt}\rightarrow \frac{\mathds{1}_{P(f(\thetab, \tilde{\xb}), \yb)} }{2} [-\lbhat_{j}(\thetab)]_+^T\left[\ubhat_{j}(\thetab)\right]_+.
\end{equation*}
\paragraph{Training details}
As common for recent certified training algorithms~\citep{Balunovic2020,Zhang2020}, $\kappa$ is linearly increased from $0$ to $1$ at the beginning of training (mixing).
Similarly to IBP~\cite{Gowal2018b} and CROWN-IBP~\citep{Zhang2020}, we also linearly increase the effective perturbation radius from $0$ to $\alpha \epsilon_{\text{ver}}$ while mixing the objectives.

\section{Verification Framework} \label{sec:verification}

As explained in $\S$\ref{sec:ibpr}, IBP-R is designed to facilitate the verification of trained networks via recent branch-and-bound frameworks. We will now first present the details of the employed complete verifier ($\S$\ref{sec:bab}), then present a novel branching strategy in $\S$\ref{sec:upb}.

\subsection{Branch–and-Bound Setup} \label{sec:bab}

\looseness=-1
Owing to its modularity and performance on large COLT-trained networks~\citep[\texttt{cifar2020} benchmark]{vnn-comp-2021}, we base our verifier on the OVAL branch-and-bound framework~\citep{BunelDP20, Bunel2020, improvedbab} from VNN-COMP-2021~\citep{vnn-comp-2021}.
Given that IBP-R explicitly seeks to minimize the area of the ReLU convex hull (see Figure~\ref{fig:relucvx}), we instantiate the framework so as to use $\beta$-CROWN~\citep{betacrown} for the bounding (see $\S$\ref{sec:background-verification}), a state-of-the-art solver for the employed relaxation, designed for use within branch-and-bound.
In line with~\citet{improvedbab}, intermediate bounds are never updated after branching, and they are individually computed via $\alpha$-CROWN~\cite{xu2021fast}. 
Furthermore, the dual variables of each bounding computation are initialized to the values associated to the parent node (that is, the bounding performed before the last split), and the number of dual iterations is dynamically adjusted to reduce the bounding time~\citep{improvedbab}.
Counter-examples are found using the MI-FGSM~\cite{dong2018boosting} adversarial attack, which is run repeatedly for each property, using a variety of hyper-parameter settings.
Verification is run with a timeout of $1800$ seconds, and terminated early when the property is likely to time out. 
We now present the employed branching strategy.

\subsection{UPB Branching} \label{sec:upb}

In spite of its strong empirical performance on COLT-trained networks~\citep{vnn-comp-2021}, the FSB branching strategy~\citep{improvedbab}, commonly employed for $\beta$-CROWN~\citep{betacrown}, requires $O(n)$ CROWN-like bounding computations per split. In order to reduce the branching overhead, we present Upper Planet Bias (UPB), a novel and simpler branching strategy that yields splitting decisions of comparable quality at the cost of a single gradient backpropagation through the~network.

The popular SR~\citep{Bunel2020} and FSB branching strategies partly rely on estimates of the sensitivity of FastLin bounds~\citep{Wong2018} to the splitting of an ambiguous ReLUs~\citep{improvedbab}.
As the employed relaxation for the output bounding is typically much tighter, these estimates are often unreliable. 
In order to improve branching performance, FSB couples these estimates with an expensive bounding step.
We propose to remove the need to compute bounds at branching time by re-using dual information from branch-and-bound's bounding step, which we perform using $\beta$-CROWN (see $\S$\ref{sec:bab}).
Specifically, we propose to score branching decisions according to a dual term associated to the bias of the upper linear constraint from the Planet relaxation (Figure~\ref{fig:relucvx}) for each ambiguous~neuron:
\begin{equation}
	\mbf{s}_{\text{UPB}, k} = 
	\hspace{-5pt} \left. \begin{array}{l}
	\frac{\left[-\lbhat_{j}\right]_+^T\left[\ubhat_{j}\right]_+ }{\left(\hat{\ub}_k-\hat{\lb}_k\right)} \odot [\bar{\lambdab}_{k}]_+ \end{array} 
	\right. \hspace{-5pt} \enskip  k \in \left\llbracket1,n-1\right\rrbracket,\\
\label{eq:intercept-score}
\end{equation}
where $[\bar{\lambdab}_{k}]_+$ is computed by evaluating equation \eqref{eq:beta-crown} in appendix~\ref{sec:betacrown} on the dual variables computed for the branch-and-bound bounding step. Therefore, the cost of computing scores \eqref{eq:intercept-score} for all neurons corresponds to that of a single gradient backpropagation.
Intuitively, as $\mbf{s}_{\text{UPB}, k}$ will disappear from the dual objective after splitting, we employ it as a proxy for a branching decision's potential bounding~improvement. See appendix~\ref{sec:betacrown} for further details.

\section{Related Work}

Many popular certified training algorithms work by upper bounding the robust loss \eqref{eq:robust-problem} via some combination of \gls{ibp} and linear bounds on the activation function. \gls{ibp}~\citep{Gowal2018b,Mirman2018}, CAP~\citep{Wong2018,Wong2018a}, and CROWN-IBP~\citep{Zhang2020} all fall in this category.
\citet{Shi2021} recently proposed a series of techniques to shorten the usually long training schedules of these algorithms. \citet{Xu2020} provide minor improvements on CROWN-IBP by changing the way the loss function is incorporated into problem \eqref{eq:relaxation}.
The above family of methods produce state-of-the-art results for larger perturbation radii. Regularization-based techniques, instead, tend to perform better on smaller radii (see $\S$\ref{sec:hybrid}).
Further works have focused on achieving robustness via specialized network architectures~\citep{Lyu2021,Zhang2021}, Lipschitz constant estimation for perturbations in the $\ell_2$ norm~\citep{Huang2021}, or under randomized settings~\citep{Cohen2019,Salman2019b}: these methods are out of the scope of the present~work.

As outlined in $\S$\ref{sec:background}, widely-employed neural network relaxations model the convex hull of the activation function, referred to as the convex barrier due to its popularity~\citep{Salman2019}, or on even looser convex outer-approximations such as FastLin~\citep{Wong2018}, CROWN~\citep{Zhang2018}, DeepZ~\citep{Singh2018}, or DeepPoly~\citep{Singh2019}.
These relaxations are relatively inexpensive yet very effective when adapted for complete verification via branch-and-bound, and are hence at the core of the $\alpha$-$\beta$-CROWN~\citep{xu2021fast,betacrown} and OVAL frameworks~\citep{BunelDP20,improvedbab}.
A number of works have recently focused on devising tighter neural network relaxations~\citep{Singh2019b,Anderson2020,Tjandraatmadja,Muller2021}. These have been integrated into recent branch-and-bound verifiers~\citep{DePalma2021,sparsealgos,Ferrari2022} and yield strong results for harder verification properties on medium-sized networks.

\looseness=-1
While the employed relaxations are a fundamental component of a complete verifier, the overall speed-accuracy trade-offs are greatly affected by the employed branching strategy. 
Small networks with few input dimensions can be quickly verified by recursively splitting the input region~\citep{Bunel2018,Royo2019}. 
On the other hand, activation splitting~\citep{Ehlers2017,Katz2017} (see $\S$\ref{sec:background-verification}, $\S$\ref{sec:upb}) performs better for larger convolutional networks~\citep{Bunel2020,improvedbab} and enjoyed recent developments based on graph neural networks~\citep{Lu2020Neural} or for use with tighter relaxations~\citep{Ferrari2022}.

\begin{table*}[t!]
    \vspace{-7pt}
	\sisetup{detect-weight=true,detect-inline-weight=math,detect-mode=true}
	
	\centering
	
	\scriptsize
	\setlength{\tabcolsep}{4pt}
	\aboverulesep = 0.1mm  
	\belowrulesep = 0.2mm  
	\begin{adjustbox}{max width=\textwidth, center}
		
		\begin{tabular}{
				c
				l
				S[separate-uncertainty,table-format=2.2(2)]
				S[separate-uncertainty,table-format=2.2(2)]
				S[separate-uncertainty,table-format=2.2(2)]
				c
			} 
			
			
			\multicolumn{1}{ c }{Perturbation} &
			\multicolumn{1}{ c }{Method} &
			\multicolumn{1}{ c }{Standard accuracy} &
			\multicolumn{1}{ c }{Robust accuracy} &
			\multicolumn{1}{ c }{Verified accuracy}  &
			\multicolumn{1}{ c }{Runtime [s]} \\
			
			\cmidrule(lr){1-1} \cmidrule(lr){2-2} \cmidrule(lr){3-5} \cmidrule(lr){6-6} \\[-5pt]
			
			\multirow{7}{*}{$\epsilon_{\text{ver}}=2/255$} &
			
			\multicolumn{1}{ c }{\textsc{COLT}} 
			& \B 78.37(24)  &  65.66(13)   &  61.88(11) & $3.05 \times 10^4 \pm 1.21 \times 10^2$ $^\ddag$ \\
			
			& \multicolumn{1}{ c }{\textsc{IBP-R}} 
			& 78.19(52) &  \B 66.39(12) & \B 61.97(18) &  $9.34 \times 10^3 \pm 2.95 \times 10^1$  \\
			
			& \multicolumn{1}{ c }{\textsc{IBP-R w/ Masking}} 
			& 78.22(26) &  66.28(17) & 61.69(29) &  $9.63 \times 10^3 \pm 4.42 \times 10^1$  \\ 
			
			\cmidrule(lr){2-6}
			
			& \multicolumn{1}{ c }{\textsc{Literature Results}} \\
			
			& \multicolumn{1}{ c }{\citep{Shi2021}$^*$} 
			& 66.84 &  /  &  52.85 &  \\
			
			& \multicolumn{1}{ c }{\citep{Zhang2020}$^*$} 
			& 71.52 &  59.72 &  53.97 & $9.13 \times 10^4$ $^\dagger$ \\ 
			
			& \multicolumn{1}{ c }{\citep{Balunovic2020}} 
			&  78.4  & / &  60.50 &  \\
			
			& \multicolumn{1}{ c }{\citep{Xiao2019}} 
			&  61.12  & 49.92 &  45.93 &  \\[3pt]
			
			\cmidrule(lr){1-6} \\[-5pt]
			
			\multirow{6}{*}{$\epsilon_{\text{ver}}=8/255$} &
			
			\multicolumn{1}{ c }{\textsc{COLT}} 
			&  51.94(14) &  31.68(23)   &  28.73(23) & $1.03 \times 10^4  \pm 1.70 \times 10^1$ $^\ddag$ \\
			
			& \multicolumn{1}{ c }{\textsc{IBP-R}} 
			& 51.43(21) &  31.89(11) &  27.87(1) & $5.92 \times 10^3  \pm 2.95 \times 10^1$ \\
			
			& \multicolumn{1}{ c }{\textsc{IBP-R w/ Masking}} 
			& 52.74(30) &  32.78(33) &  27.55(22) & $5.89 \times 10^3  \pm 3.38 \times 10^1$  \\ 
			
			\cmidrule(lr){2-6}
			
			& \multicolumn{1}{ c }{\textsc{Literature Results}} \\
			
			& \multicolumn{1}{ c }{\citep{Shi2021}$^*$} 
			& 48.28(40) &  /  &  \B 34.42(32) &  $9.51 \times 10^3$ $^\diamond$ \\ 
			
			& \multicolumn{1}{ c }{\citep{Zhang2020}$^*$} 
			& \B 54.50 &  \B 34.26 &  30.50 & $9.13 \times 10^4$ $^\dagger$ \\ 
			
			& \multicolumn{1}{ c }{\citep{Balunovic2020}} 
			&  51.70  & / &  27.50 &  \\
			
			& \multicolumn{1}{ c }{\citep{Xiao2019}} 
			&  40.45  & 26.78 &  20.27 &  \\
			
			\bottomrule 
			
			\\[-5pt]
			\multicolumn{6}{ c }{\scriptsize
				\begin{minipage}{.9\textwidth}
					$^*$ the employed $7$-layer network has $17.2 \times 10^6$ parameters, as opposed to the $2.1 \times 10^6$ parameters of the $5$-layer and $4$-layer networks respectively used for our $\epsilon_{\text{ver}}=2/255$ and $\epsilon_{\text{ver}}=8/255$ experiments. \\
					$^\diamond$ the training is performed on an Nvidia RTX 2080 Ti GPU, which is significantly faster than that employed in our experiments. \\
					$^\dagger$ the training is performed on $4$ Nvidia RTX 2080 Ti GPUs. \\
					$^\ddag$ on the same setup, the original PyTorch implementation runs in $1.74 \times 10^5$ seconds for $\epsilon_{\text{ver}}=2/255$, and $4.49 \times 10^4$ seconds for $\epsilon_{\text{ver}}=8/255$. 
				\end{minipage}
			}

		\end{tabular}
	\end{adjustbox}
	\vspace{-3pt}
	\caption{\small Performance of different verifiably robust training algorithms under $\ell_\infty$ norm perturbations, on the CIFAR10 dataset with $\epsilon_{\text{ver}}=2/255$ and $\epsilon_{\text{ver}}=8/255$. 
		The table reports mean and sample standard deviation over three repetitions for our experiments, over five repetitions for \citep{Shi2021} on $\epsilon_{\text{ver}}=8/255$. The remaining results from the literature were executed with a single seed. 
		The method with the best average performance for each perturbation radius is highlighted in bold. 
		\label{fig:cifar-tables}}
	\vspace{-5pt}
\end{table*}

\section{Experiments} \label{sec:experiments}

\looseness=-1
In this section, we present an experimental evaluation of the IBP-R certified training algorithm ($\S$\ref{sec:experiments-train}), then evaluate the performance of our UPB branching strategy ($\S$\ref{sec:experiments-verify}).
The implementation of our training algorithm is based on Jax~\citep{jax2018github}, while verification is performed post-training by using a modified version (see~$\S$\ref{sec:verification}) of the OVAL framework, implemented in PyTorch~\citep{Paszke2019}.
Code is available at \url{https://github.com/alessandrodepalma/ibpr}.

\subsection{Verified Training} \label{sec:experiments-train}

We evaluate the efficacy of our IBP-R algorithm~($\S$\ref{sec:ibpr}) by replicating the CIFAR-10~\citep{CIFAR10} experiments from~\citet{Balunovic2020} and comparing against COLT, which we ported to Jax for fairness (resulting in significant speed-ups as shown in table~\ref{fig:cifar-tables}, see appendix~\ref{sec:jax-colt} for details). 
We focus our comparison on COLT, as it is the best-performing instance of the hybrid training algorithms detailed in $\S$\ref{sec:hybrid}, family to which IBP-R belongs. 
Furthermore, to the best of our knowledge, it yields state-of-the-art results for small perturbations and ReLU networks.
Details concerning the employed architecture, hyper-parameters, and the computational setup can be found in appendix~\ref{sec:exp-appendix}.
Timing results were executed on a Nvidia Titan V GPU.

Table~\ref{fig:cifar-tables} reports the results of our experiment, as well as relevant results from the literature. Specifically, we report results for: 
(i) CROWN-IBP~\citep{Zhang2020} and the improved IBP algorithm by \citet{Lyu2021}, representing state-of-the-art certified training algorithms based on upper bounding the robust loss (see $\S$\ref{sec:robust-loss}),
(ii) \citet{Xiao2019}, sharing many similarities with IBP-R (see $\S$\ref{sec:hybrid}),
and (iii) the original COLT experiments from \citet{Balunovic2020}.
IBP-R excels on the smaller $\ell_\infty$ perturbation radius ($\epsilon_{\text{ver}}=2/255$), displaying verified and standard accuracies comparable (given the experimental variability) to COLT, and a larger empirically robust accuracy (for details on the employed attack, see $\S$\ref{sec:bab}). 
Furthermore, IBP-R training is more than three times faster than COLT, making its use particularly convenient on this setup.
The masking (see $\S$\ref{sec:ibpr}) does not appear to be beneficial on smaller perturbations.
Relevant results from the literature all underperform compared to IBP-R, which hence achieves state-of-the-art results on this benchmark.
\looseness=-1
On the larger perturbation radius ($\epsilon_{\text{ver}}=8/255$), masking the regularization has a positive effect.
The masked version of IBP-R performs comparably with COLT in $\approx 57\%$ of its runtime, attaining larger standard and empirically robust accuracies, and smaller verified accuracy.
However, in this context, all regularization-based methods (including IBP-R) are outperformed by algorithms upper bounding the robust loss~\citep{Zhang2020,Lyu2021}. Nevertheless, these works employ significantly larger networks than the one used in our experiments. We therefore conjecture that IBP-R, which scales better than COLT with the network depth (see runtimes in table~\ref{fig:cifar-tables}), will become more competitive when evaluated on comparable settings.

\subsection{Branching} \label{sec:experiments-verify}

In order to test the efficacy of our UPB branching strategy~($\S$\ref{sec:upb}), we time verification for the first $500$ CIFAR-10 test samples on two IBP-R-trained networks (one per perturbation radius). 
Images that are incorrectly classified are discarded.
We keep the branch-and-bound settings fixed to those of $\S$\ref{sec:bab}, and benchmark against the following branching strategies: FSB~\citep{improvedbab}, and SR~\citep{Bunel2020}.
Appendix~\ref{sec:supp-branching} replicates the experiment on two COLT-trained networks.

\begin{figure}[b!]
    \vspace{-10pt}
	\begin{subtable}{\columnwidth}
		\sisetup{detect-weight=true,detect-inline-weight=math,detect-mode=true}
		\centering
		
		\begin{adjustbox}{max width=\textwidth, center}
			\begin{tabular}{
					l
					S[table-format=4.3]
					S[table-format=4.3]
					S[table-format=4.3]
					S[table-format=4.3]
					S[table-format=5.3]
					S[table-format=4.3]
				}
				{} & \multicolumn{3}{l}{$\epsilon_{\text{ver}}=2/255$ IBP-R} & \multicolumn{3}{l}{$\epsilon_{\text{ver}}=8/255$ IBP-R w/ Masking} \\
				\toprule
				\multicolumn{1}{ c }{Method} &
				\multicolumn{1}{ c }{time [s]} &
				\multicolumn{1}{ c }{subproblems$^*$} &
				\multicolumn{1}{ c }{$\%$Timeout} &
				\multicolumn{1}{ c }{time [s]} &
				\multicolumn{1}{ c }{subproblems$^*$} &
				\multicolumn{1}{ c }{$\%$Timeout} \\
				\midrule
				\textsc{UPB} & \B 113.42 &  \B     930.83 &  \B   5.70 &  \B 237.24 &       935.80 &   \B 12.55 \\
				\textsc{FSB} &   127.65 &      1244.01 &     6.22 &   243.29 &       \B 795.22 &    12.92 \\
				\textsc{SR} &   215.89 &      5185.27 &    11.66 &   303.76 &      6744.72 &    15.13 \\
				\bottomrule
				
				\\[-8pt]
				\multicolumn{7}{ c }{\scriptsize
					\begin{minipage}{1.5\columnwidth}
						$^*$computed on the properties that did not time out for neither UPB nor FSB. The inclusion of timed-out results in the average leads to an overestimation of the number of subproblems for the less expensive branching strategy.
					\end{minipage}
				}
				
			\end{tabular}
		\end{adjustbox}
		\vspace{-3pt}
		\subcaption{\label{table:branching-ibpr}\small Comparison of average runtime, average number of solved subproblems and the percentage of timed out properties. The best performing method is highlighted in bold.}
	\end{subtable}
	\begin{subfigure}{\columnwidth}
		\begin{subfigure}{.49\textwidth}
			\centering
			\includegraphics[width=\textwidth]{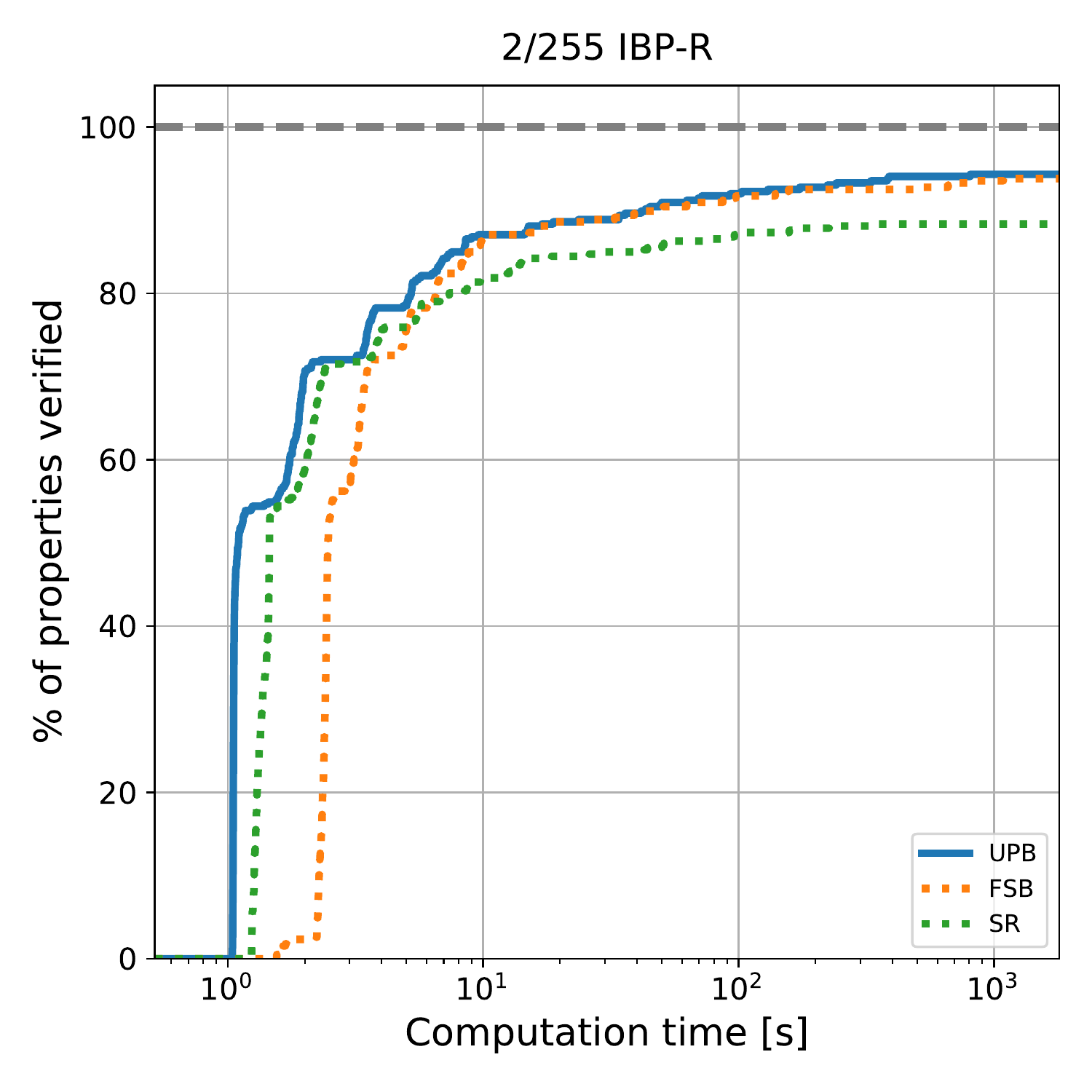}
		\end{subfigure}
		\begin{subfigure}{.49\textwidth}
			\centering
			\includegraphics[width=\textwidth]{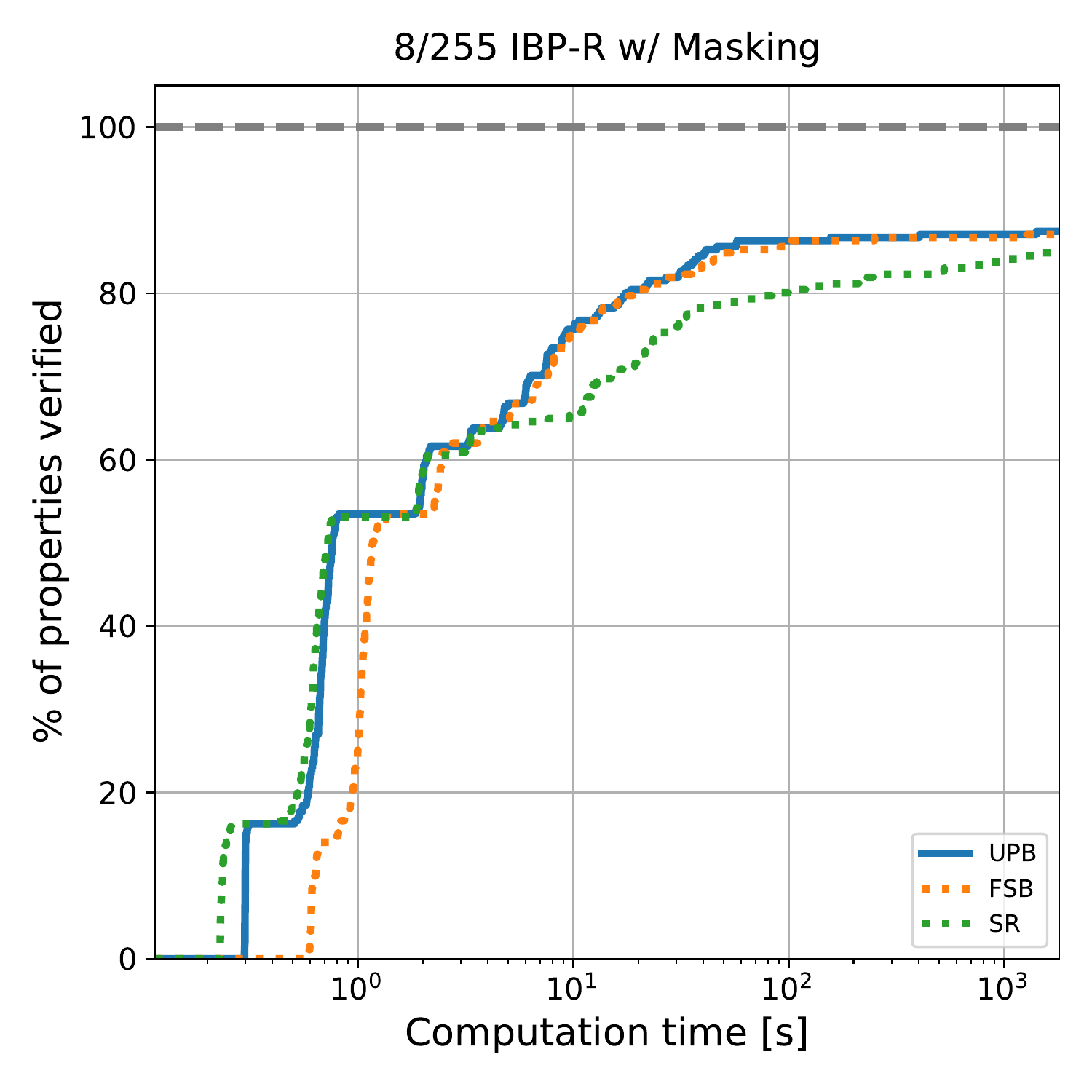}
		\end{subfigure}
		\vspace{-7pt}
		\subcaption{Cactus plots: percentage of solved properties as a function of runtime. Baselines are represented by dotted lines.}
	\end{subfigure}
	\vspace{-5pt}
	\caption{\label{fig:branching-ibpr} Complete verification performance of different branching strategies on two IBP-R-trained CIFAR-10 networks. }
\end{figure}

\looseness=-1
Figure~\ref{fig:branching-ibpr} shows that UPB improves average verification times compared to FSB (by roughly $13\%$ and $2.5\%$ for the $\epsilon_{\text{ver}}=2/255$ and $\epsilon_{\text{ver}}=8/255$ networks, respectively). 
On the larger perturbation radius, this holds in spite of a larger average number of visited subproblems, highlighting the cost of FSB's bounding-based selection step. 
On the other hand, UPB verifies more properties within the timeout on both networks. Furthermore, it reduces the number of visited subproblems on the smaller perturbation, testifying the efficacy of the selected domain splits. 
Both UPB and FSB yield significantly faster verification than SR on the considered problems.
The results show that UPB is less expensive than the state-of-the-art FSB algorithm, while producing branching decisions of comparable quality.

\section{Conclusions}

Many state-of-the-art verified training algorithms require very large networks to obtain good robustness-accuracy trade-offs.
Methods designed to exploit tight bounds at verification-time better exploit network capacity but they either underperform or involve extremely complex training procedures.
We introduced IBP-R, a simple and intuitive robust training algorithm designed to induce verifiability via recent branch-and-bound algorithms. We show that, by minimizing the area of the Planet relaxation via IBP bounds for all network activations, and using PGD attacks over larger perturbations, one can obtain state-of-the-art certified accuracy results on small perturbations without large compromises in standard accuracy.
Finally, in order to ease the task of verifying the trained networks, we presented UPB, a straightforward and inexpensive branching strategy that yields branching decisions as effective as the state-of-the-art.
We believe our results could be further improved by leveraging recent improvements on standard IBP training~\citep{Shi2021}, as well as larger network architectures: these are interesting avenues for future work.

%
%


\bibliography{ibpreg}
\bibliographystyle{wfvml2022}

\newpage
\appendix
\onecolumn

\section{Comparison between IBP-R and COLT} \label{sec:colt-comparison}

In this section, we provide a detailed comparison between COLT by \citet{Balunovic2020} ($\S$\ref{sec:hybrid}) and IBP-R ($\S$\ref{sec:ibpr}) in the context of training for adversarial robustness. 
Equation \eqref{eq:colt-diff} pictorially highlights the differences between the two algorithms, which are summarized in table~\ref{tab:colt-diff}.

\begin{table}[h!]
	\centering \small
	\begin{adjustbox}{max width=\textwidth, center}
		\begin{tabular}{ccc}
			\textbf{Training Detail}\TBstrut & \textbf{COLT}\TBstrut & \textbf{IBP-R} \TBstrut\\
			\toprule 
			\vspace{-5pt} &&\\
			Stages
			&
			\begin{minipage}{.4\columnwidth}
			Proceeds in a stage-wise fashion: at each stage, a subset of the network parameters (all the $\thetab$ that are not in $\thetab^j$) is frozen, and the remaining parameters are optimized over. The number of training stages is $O(n)$.
			\end{minipage}
			& \begin{minipage}{.4\columnwidth}
			All network parameters are optimized at once.
			\end{minipage} \vspace{7pt} \\
			PGD domains &
			\begin{minipage}{.4\columnwidth}
			The attacks are carried out in the space of the $j$-th activations ($\xb_j$).
			The frozen subset of the network is replaced by a zonotope-based convex outer-approximation, employed to define $\mathbb{C}_j(\xb)$ and $\mathbb{C}_{j-1}(\xb)$, the domains of the PGD attacks. 
			At each stage, the zonotope domains are computed for train-time perturbations larger than those to be verified: $\epsilon_{\text{train}, j} > \epsilon_{\text{ver}}$.
			In practice, the CIFAR-10 experiments from \citet{Balunovic2020} use $1.05 \leq \epsilon_{\text{train}, j} \leq 1.512$.
			For the first stage: $\mathbb{C}_0(\xb) := \mathcal{C}_+(\xb)$, $\mathbb{C}_{-1}(\xb) := \xb$.
			\end{minipage}
			& \begin{minipage}{.4\columnwidth}
			The attacks are performed in the network input space, over a superset of the perturbations employed at verification. Our notation for $\mathcal{C}_+(\xb)$ corresponds to setting
			$\epsilon_{\text{train}} := \alpha \epsilon_{\text{ver}}$, with $1.6 \leq \alpha \leq 2.1$ in our experiments (after the mixing phase).
			Note that the first stage of COLT displays the same attack structure, with smaller employed $\alpha$ values.
			\end{minipage} \vspace{7pt} \\
			Intermediate bounds & \begin{minipage}{.4\columnwidth}
			Computed via an approximation of the zonotope relaxation based on Cauchy random projections~\citep{Li2007}. These bounds, computed for perturbation radius $\epsilon_{\text{train}, j}$, are used both for the regularization term and to define the zonotope outer-approximations (which depend on intermediate bounds).
			\end{minipage} & \begin{minipage}{.4\columnwidth}
			Computed via \gls{ibp}, for perturbation radius $\epsilon_{\text{train}}$. Used for the regularization term.
			\end{minipage}\vspace{7pt} \\
			Bounds regularization & \begin{minipage}{.4\columnwidth}
			Minimizes the area of the zonotope relaxation of ambiguous ReLUs for a single later per stage. This term produces a non-null gradient only for the parameters of the $(j+1)$-th layer.
			\end{minipage} & \begin{minipage}{.4\columnwidth}
			Minimizes the area of the Planet relaxation (in practice, one half of the area of the zonotope relaxation) of ambiguous ReLUs for all layers at once (including the output space $\hat{x}_n$) .
			\end{minipage}\vspace{7pt} \\
			Hyper-parameters & \begin{minipage}{.4\columnwidth}
			Each stage is potentially associated to a different regularization coefficient $\rho_j$, to a different train-time perturbation radius $\epsilon_{\text{train}, j}$, and to a different learning rate $\eta_j$. In practice, \citet{Balunovic2020} tune the values for the first stage ($\rho_0$, $\epsilon_{\text{train}, 0}$, $\eta_0$), and then respectively decay $\rho_0$ and $\epsilon_{\text{train}, 0}$, and increase $\rho_0$, by a fixed and tunable quantity at each stage.
			Finally, \citet{Balunovic2020} set $\rho_{n-2}=0$ in all cases.
			\end{minipage} & \begin{minipage}{.4\columnwidth}
			$\rho$, $\alpha$, and $\eta$ are not altered throughout training.
			\end{minipage} \vspace{7pt} \\
			Mixing phase & \begin{minipage}{.4\columnwidth}
			At each stage, $\kappa$ is linearly increased from $0$ to $1$.
			\end{minipage} & \begin{minipage}{.4\columnwidth}
			$\kappa$ is linearly increased from $0$ to $1$, $\epsilon_{\text{train}}$ is linearly increased from $0$ to $\alpha \epsilon_{\text{ver}}$.
			\end{minipage} \\
			\vspace{-5pt} &&\\
			\bottomrule
		\end{tabular}
	\end{adjustbox}
	\caption{\label{tab:colt-diff} Main differences between the COLT ($\S$\ref{sec:hybrid}) and IBP-R ($\S$\ref{sec:ibpr}) verified training algorithms.}
\end{table}

\begin{equation}
    \begin{aligned}
    \text{COLT:} \hspace{200pt}& \text{\hspace{20pt}IBP-R:} \\
	\colorlet{savedleftcolor}{.}\color{purple}\left[\color{savedleftcolor} \hspace{-5pt} \begin{array}{l} \min_{\color{purple}\thetab^{j}} \EX_{(\xb, \yb) \in \mathcal{D}} 
	\left\{ \hspace{-5pt} \begin{array}{l}
	 \kappa \left[\begin{array}{l} \max_{\color{purple}\xb' \in \mathbb{C}_j(\xb)} \mathcal{L} (\textcolor{purple}{f^{j}}(\thetab^{j}, \xb'), \yb) \\ 
	 + \textcolor{purple}{\rho_j} \left[\textcolor{purple}{-\lbhat_{j+1}(\thetab^j)}\right]_+^T\left[\textcolor{purple}{\ubhat_{j+1}(\thetab^j)}\right]_+ \end{array}\right] \\[15pt]
	 + (1 - \kappa) \max\limits_{\color{purple}\xb' \in \mathbb{C}_{j-1}(\xb)} \mathcal{L} (\textcolor{purple}{f^{j-1}}(\thetab^{j-1}, \xb'), \yb)\\ 
	 + \lambda \left\lVert \thetab^j \right\rVert_1, \end{array}\right. \end{array} \hspace{-10pt} \color{purple}\right]
	 \qquad
	 &\min_{\color{teal}\thetab} \hspace{-2pt} \EX_{(\xb, \yb) \in \mathcal{D}} 
	 \hspace{-2pt} \left\{ \hspace{-5pt} \begin{array}{l}
		\kappa \left[\begin{array}{l}\max_{\color{teal}\xb \in \mathcal{C}_+(\xb)} \mathcal{L} (\textcolor{teal}{f}(\thetab, \xb), \yb)\ + \\ \textcolor{teal}{\frac{\rho}{2} \sum_{j=1}^{n}} \left[\textcolor{teal}{-\lbhat_{j}(\thetab)}\right]_+^T\left[\textcolor{teal}{\ubhat_{j}(\thetab)}\right]_+ \end{array}  \right] \\[15pt]
		+\ (1 - \kappa) \mathcal{L} (\textcolor{teal}{f}(\thetab, \xb), \yb) + \lambda \left\lVert \thetab \right\rVert_1. \end{array}\right. \\
		\textcolor{purple}{\forall j \in \left\llbracket0,n-2\right\rrbracket} \hspace{5pt} &
	   \end{aligned}
	   \label{eq:colt-diff}
\end{equation}

\section{Complete Verification Problem} \label{sec:complete-verification}

Provided the network is in canonical form~\citep{Bunel2020}, complete verification amounts to finding sign of the minimum of the following problem, of which problem \eqref{eq:relaxation} is a convex outer-approximation:
\begin{equation}
	\begin{aligned}
		&\smash{\min_{\xb, \xbhat}}\quad \hat{x}_{n} \qquad \text{s.t. } \\[7pt] 
		& \xb_0 \in \mathcal{C}(\xb), \\
		& \xbhat_{k+1} = W_{k+1} \xb_{k} + \mathbf{b}_{k+1} && k \in \left\llbracket0,n-1\right\rrbracket,\\
		& \xb_k = \sigma(\xbhat_{k}) && k \in \left\llbracket1,n-1\right\rrbracket.
	\end{aligned}
	\label{eq:complete-verification}
\end{equation}

\section{Intermediate Bounds} \label{sec:ibs}

As seen in $\S$\ref{sec:background-verification}, the network relaxations employed for incomplete verification (and, hence, complete verification via branch-and-bound) are defined in terms of bounds on the network pre-activations (intermediate bounds $\hat{\lb}_k, \hat{\ub}_k$).
Intermediate bounds are computed by solving instances of problem \eqref{eq:relaxation} over subsets of the network. 
For instance, for  $\hat{\lb}_i[j]$, the lower bound on $\xbhat_i[j]$:
\begin{equation}
	\begin{aligned}
		\min_{\xb, \xbhat}  \enskip &\xbhat_{i}[j]  \\
		& \xb_0 \in \mathcal{C}(\xb), \\
		& \xbhat_{k+1} = W_{k+1} \xb_{k} + \mathbf{b}_{k+1} && k \in \left\llbracket0,i-1\right\rrbracket,\\
		& (\xb_k, \xbhat_{k}) \in \text{Rel}(\sigma, \hat{\lb}_k, \hat{\ub}_k) && k \in \left\llbracket1,i-1\right\rrbracket, \\
		& \xbhat_k \in [\hat{\lb}_k, \hat{\ub}_k]   && k \in \left\llbracket1,i-1\right\rrbracket,
	\end{aligned}
	\label{eq:primal-intermediates}
\end{equation}
where intermediate bounds until the $(i-1)$-th layer are needed.
Overall, problem \eqref{eq:primal-intermediates} needs to be solved twice per neuron: once for the lower bound, once for the upper bound, by flipping the sign of the last layer's weights.
Therefore, intermediate bounds are often computed by relying on looser relaxations than for the output bounding (that is, solving for $\smash{\min_{\xb, \xbhat}}\enskip \hat{x}_{n}$).
See $\S$\ref{sec:bab} for details on how we compute intermediate bounds within branch-and-bound.

\section{Interval Bound Propagation} \label{sec:ibp}

\begin{figure}[t!]
	\centering
	\noindent\resizebox{.28\textwidth}{!}{
		\definecolor{lightgray}{rgb}{0.8,0.8,0.8}

\begin{tikzpicture}
  \tikzset{dummy/.style= {inner sep=0, outer sep=0}}
  \tikzset{cross/.style={circle, draw,
      minimum size=5*(#1-\pgflinewidth),
      inner sep=0pt, outer sep=0pt,
      ultra thick, color=red}}


  \draw[dashed](-1, -0.5) to (-1, 1.5);
  \draw[dashed](1, -0.5) to (1, 1.5);
  \draw[dashed](-1.5, 1) to (2, 1);

  \draw[fill=teal!80, fill opacity=0.9](-1, 0) -- (1,0) -- (1, 1) -- (-1, 1) -- (-1, 0);

  \node[dummy](lb-lab) at (-1.3, -0.3) {$\hat{l}_{k[j]}$\hspace{3pt}};
  \node[dummy](ub-lab) at (1.35, -0.3) {\hspace{3pt}$\hat{u}_{k[j]}$};
  \node[dummy](ub-lab) at (2.2, 1.25) {$\xb_k[j]=\hat{u}_{k[j]}$};


  \draw[-latex](-1.5,0) to (2, 0);
  \node[dummy](x-label) at (2.3, 0) {\hspace{3pt} $\xbhat_{k[j]}$};
  \draw[-latex](0,-0.5) to (0, 1.5);
  \node[dummy](x-label) at (0, 1.8) {$\xb_{k[j]}$};

\end{tikzpicture}
	}
	\caption{Depiction of the \gls{ibp} hyper-rectangle.}
	\label{fig:ibp}
\end{figure}
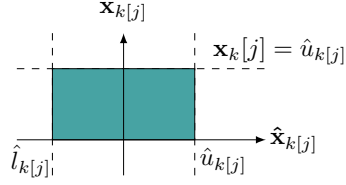

Interval bound propagation~\citep{Gowal2018b, Mirman2018}, a simple application of interval arithmetic~\citep{Sunaga1958,Hickey2001} to neural networks,
implies solving a version of problem \eqref{eq:relaxation} where $\text{Rel}(\sigma, \hat{\lb}_k, \hat{\ub}_k)$ is the hyper-rectangle depicted in Figure~\ref{fig:ibp}. We will use $\mathcal{B}$ to denote the corresponding feasible region.
Furthermore, let us write $[\xb]_- := \min(\xb, \mbf{0})$ and $\hat{\lb}_{n} := \smash{\min_{\xb, \xbhat}}\enskip \hat{x}_{n} \text{ s.t.} \left(\xb, \xbhat\right) \in \mathcal{B}$.
Due to the simplicity of the relaxation, the problem enjoys the following closed form solution:
\begin{equation*}
		\hat{\lb}_{n} := \smash{\min_{\xb, \xbhat}}\enskip \hat{x}_{n}  =  \left[W_{n}\right]_+ [\hat{\lb}_{n-1}]_+ + [W_{n}]_- [\hat{\ub}_{n-1}]_+ + \bb_{n}.
\end{equation*}
Upper bounds can be computed by replacing $\hat{\lb}_{n-1}$ with $\hat{\ub}_{n-1}$, and viceversa.
Note that the bounds at layer $n$ only depend on the intermediate bounds at layer $(n-1)$. Therefore, output bounds and all intermediate bounds can be computed at once by forward-propagating the bounds from the first layers, at the total cost of four network evaluations. 
Let $d$ be the input dimensionality: $\xb_{0} \in \mathbb{R}^d$.
\gls{ibp} is significantly less expensive than relaxations based on linear bounds~\citep{Wong2018,Zhang2018}, which incur a cost equivalent to $O(d)$ network evaluations for the same computation~\citep{Xu2020}.

\section{UPB Branching: Beta-CROWN Dual and Planet relaxation} \label{sec:betacrown}

Our UPB (see $\S$\ref{sec:upb}) branching strategy makes use of variables from the Beta-CROWN~\citep{betacrown} dual objective. We will now describe the objective as well as its relationship to the Planet relaxation (see $\S$\ref{sec:background-verification}).

\subsection{Planet Relaxation}
Let us denote by $\text{Conv}(\sigma, \hat{\lb}_k, \hat{\ub}_k)$ the element-wise convex hull of the activation function, as a function of intermediate bounds. For ReLUs, this corresponds to the Planet relaxation, which is depicted in Figure~\ref{fig:relucvx} for the ambiguous case. 
By replacing $\text{Rel}(\sigma, \hat{\lb}_k, \hat{\ub}_k)$ with $\text{Conv}(\sigma, \hat{\lb}_k, \hat{\ub}_k)$ in problem \eqref{eq:relaxation}, we obtain:
\begin{equation}
	\begin{aligned}
		&\smash{\min_{\xb, \xbhat}}\quad \hat{x}_{n} \qquad \text{s.t. } \\[7pt] 
		& \xb_0 \in \mathcal{C}(\xb), \\
		& \xbhat_{k+1} = W_{k+1} \xb_{k} + \mathbf{b}_{k+1} && k \in \left\llbracket0,n-1\right\rrbracket,\\
		& (\xb_k, \xbhat_{k}) \in \text{Conv}(\sigma, \hat{\lb}_k, \hat{\ub}_k) && k \in \left\llbracket1,n-1\right\rrbracket, \\
		& \xbhat_k \in [\hat{\lb}_k, \hat{\ub}_k]   && k \in \left\llbracket1,n-1\right\rrbracket.
	\end{aligned}
	\label{eq:primal-planet}
\end{equation}
Problem \eqref{eq:primal-planet} can be alternatively written as:
\begin{equation}
	\begin{aligned}
		\smash{\max_{\balpha \in [\mbf{0}, \mbf{1}]} \min_{\xb, \xbhat}}\quad \hat{x}_{n} \qquad
		\text{s.t. }\quad& \xb_0 \in \mathcal{C}, \\
		& \xbhat_{k+1} = W_{k+1} \xb_{k} + \mathbf{b}_{k+1} && k \in \left\llbracket0,n-1\right\rrbracket,\\
		& \ubar{\mbf{a}}_k (\balpha_k) \odot \xbhat_{k} + \ubar{\mbf{b}}_k \leq \xb_{k} \leq \bar{\mbf{a}}_k \odot \xbhat_{k} + \bar{\mbf{b}}_k && k \in \left\llbracket1,n-1\right\rrbracket, \\
		& \xbhat_k \in [\hat{\lb}_k, \hat{\ub}_k]  && k \in \left\llbracket1,n-1\right\rrbracket,
	\end{aligned}
	\label{eq:primal-betacrown}
\end{equation}
where the coefficients are defined as follows:
\begin{equation}
	\begin{aligned}
		\ubar{\mbf{a}}_k(\balpha_k)=\balpha_k,\quad \bar{\mbf{a}}_k=\frac{\hat{\ub}_k}{\hat{\ub}_k - \hat{\lb}_k}, \quad \bar{\mbf{b}}_k=\frac{- \hat{\lb}_k \odot \hat{\ub}_k}{\hat{\ub}_k - \hat{\lb}_k}  & \qquad \text{if } \hat{\lb}_k \leq  0 \text{ and } \hat{\ub}_k \geq 0, \\[5pt]
		\bar{\mbf{a}}_k=\ubar{\mbf{a}}_k(\balpha_k)=0 & \qquad  \text{if }  \hat{\ub}_k \leq 0, \\[5pt]
		\bar{\mbf{a}}_k=\ubar{\mbf{a}}_k=1 & \qquad   \text{if } \hat{\lb}_k \geq 0, \\[5pt]
		\ubar{\mbf{b}}_k=0 & \qquad   \text{in all cases}.
	\end{aligned}
	\label{eq:slopes-biases}
\end{equation}
For ambiguous ReLUs, $\bar{\mbf{a}}_k \odot \xbhat_{k} + \bar{\mbf{b}}_k$, represents the upper bounding constraint of Figure~\ref{fig:relucvx}. The bias of such constraint, $\bar{\mbf{b}}_k=\frac{- \hat{\lb}_k \odot \hat{\ub}_k}{\hat{\ub}_k - \hat{\lb}_k}$, gives the name to our branching strategy (Upper Planet Bias).

\subsection{Beta-CROWN Dual}
Within branch-and-bound for ReLU networks, the constraints of the form $\xbhat_k \in [\hat{\lb}_k, \hat{\ub}_k]$ can be usually omitted from \eqref{eq:primal-betacrown}, except when they capture the additional constraints associated to the domain splits (split constraints).
For simplicity, we will enforce split constraints on all stable neurons, regardless of whether stability comes from actual split constraints or held before splitting.
In this context, using \citep[equations (8), (9), (38)]{Salman2019}, the Lagrangian relaxation of problem \eqref{eq:primal-betacrown} can be written as follows:  
\begin{equation}
	\begin{aligned}
		\max_{\balpha \in [\mbf{0}, \mbf{1}], \mub,\lambdab, \beta} & \min_{\xb, \xbhat} \left[ \begin{array}{l}
			W_n\xb_{n-1} + b_n + \sum_{k=1}^{n-1} \mub_k^T \left( \xbhat_k - W_k \xb_{k-1} - \mbf{b}_k \right) \\[4pt]
			+\sum_{k=1}^{n-1} [\lambdab_k]_-^T \left( \xb_k - (\ubar{\mbf{a}}_k (\balpha_k) \odot \xbhat_{k} + \ubar{\mbf{b}}_k) \right) + \sum_{k=1}^{n-1} [\lambdab_k]_+^T \left( \xb_k - (\bar{\mbf{a}}_k \odot \xbhat_{k} + \bar{\mbf{b}}_k) \right) \\[4pt]
			+ \sum_{k=1}^{n-1} \bbeta_k^T \xb_{k} \mathds{1}_{\hat{\ub}_k \leq 0} - \sum_{k=1}^{n-1} \bbeta_k^T \xb_{k} \mathds{1}_{\hat{\lb}_k \geq 0}
		 \end{array} \right.\\
		\hspace{-5pt}\text{s.t. }\enskip& \xb_0 \in \mathcal{C}.
	\end{aligned}
	\label{eq:dual-propagation}
\end{equation}
By enforcing the coefficient of the unconstrained $\xb$ and $\xbhat$ terms to be null, we obtain the following: 
\begin{equation}
\begin{aligned}
\hspace{-5pt}d_P &=  \max_{\balpha \in [\mbf{0}, \mbf{1}],\ \bbeta \geq 0} \left\{\min_{\xb_0 \in \mathcal{C}} 
\left(- \bar{\mub}_{1}^T W_1 \xb_{0}\right) +  b_n
- \sum_{k=1}^{n-1} \left( [\bar{\lambdab}_k]_-^T  \ubar{\mbf{b}}_k + [\bar{\lambdab}_k]_+^T\bar{\mbf{b}}_k +  \bar{\mub}_k^T\mbf{b}_k\right)\right\} \\
\hspace{-5pt}\text{s.t. }\qquad& \bar{\lambdab}_{n-1} = -W_{n}^T,\\
&\bar{\mub}_k = \bar{\mbf{a}}_k \odot [\bar{\lambdab}_k]_+ +  \balpha_k \odot [\bar{\lambdab}_k]_- + \mbf{s}_k \odot \bbeta_k \qquad k \in \left\llbracket1, n-1\right\rrbracket,\\
&\bar{\lambdab}_{k-1} = W_k^T\bar{\mub}_k \hspace{85pt} k \in \left\llbracket2, n-1\right\rrbracket, \\[10pt]
\text{where: }\qquad& \mbf{s}_k = \mathds{1}_{\hat{\lb}_k \geq 0} - \mathds{1}_{\hat{\ub}_k \leq 0}.
\end{aligned}
\label{eq:beta-crown}
\end{equation}
Problem \eqref{eq:beta-crown} corresponds to the $\beta$-CROWN objective, as it can be easily seen by comparing it with \citep[equation (20)]{betacrown} and pointing out that, in their formulation, the input domain represents $\ell_\infty$ norm perturbations of radius $\epsilon$.
The dual variables $\bbeta$, which give the name to the algorithm, are necessary only for the neurons whose domains have been split within branch-and-bound~\citep{betacrown}. 

\subsection{UPB Branching}
The key observation behind our UPB branching strategy is that the $[\bar{\lambdab}_k]_+^T\bar{\mbf{b}}_k$ term is present only for ambiguous neurons (see the coefficients in equation \eqref{eq:slopes-biases}, note that $[\bar{\lambdab}_k]_-^T  \ubar{\mbf{b}}_k=0$).
We can hence heuristically employ it as a proxy for the improvement that a split constraint will have on $d_P$ from \eqref{eq:beta-crown}.
Replacing $\ubar{\mbf{b}}_k$ with its definition in $[\bar{\lambdab}_k]_+^T\bar{\mbf{b}}_k$ yields the branching scores $\mbf{s}_{\text{UPB}, k}$ in equation \eqref{eq:intercept-score}.
Given the values for $\balpha$ and $\bbeta$ obtained in the last bounding step within branch-and-bound, $[\bar{\lambdab}_k]_+$, and hence the scores, can be computed using their definition in problem \eqref{eq:beta-crown}, which has a cost equivalent to a single gradient backpropagation through the network. 
The UPB branching strategy then proceeds by enforcing split constraints on the neuron associated to the largest $\mbf{s}_{\text{UPB}, k}$ score throughout the network.

\section{Experimental Details} \label{sec:exp-appendix}

We now present experimental details that were omitted from $\S$\ref{sec:experiments}.
In particular, we describe the computational setup, network architectures, employed hyper-parameters, and details concerning the Jax porting of the COLT algorithm~\citep{Balunovic2020}.

\subsection{Experimental Setting, Hyper-parameters}

All the experiments were run on a single GPU, either an Nvidia Titan Xp, or an Nvidia Titan V. 
The timing experiments for verified training were all run on an Nvidia Titan V GPU, on a machine with a $20$-core Intel i9-7900X CPU.
The branching experiments were instead consistently run on an Nvidia Titan XP GPU, on a machine with a $12$-core Intel i7-6850K CPU.

\paragraph{Training hyper-parameters and additional details}
We train on the full CIFAR-10 training set, and evaluate on the full test set. 
COLT was run with the hyper-parameters provided by the authors~\citep{Balunovic2020}, whereas the hyper-parameters for IBP-R are listed in Figure~\ref{tab:hyperparameters}. 
Note that the learning rate is annealed, at each epoch, only after the mixing phase of training. Please refer to equation \eqref{eq:ibpr} for the meaning of the various hyper-parameters. 
We did not tune the parameters of the PGD attacks, which were set as for COLT: we report them for convenience. 
IBP-R hyper-parameters were directly tuned on the evaluation set as common in previous work~\citep{Gowal2018b,Zhang2020,Shi2021}.
Complying with the COLT implementation\footnote{https://github.com/eth-sri/colt/}, we normalize inputs to the networks and use data augmentation (random horizontal flips and croppings) on CIFAR-10. 

\paragraph{Verification hyper-parameters}
We now complement section $\S$\ref{sec:bab} with omitted details concerning the configuration of the OVAL branch-and-bound framework~\citep{BunelDP20, Bunel2020, improvedbab}. These details apply to both the training and the branching experiments.
The entire verification procedure is run on a single GPU. For the UPB and SR branching strategies, the branching and bounding steps are performed in parallel for batches of $600$ and $1200$ subproblems, respectively. For the FSB branching strategy, these numbers were reduced to $500$ and $1000$ subproblems, respectively, in order to prevent PyTorch out-of-memory errors.
$\beta$-CROWN is run with a dynamically adjusted number of iterations (see \citep[section 5.1.2]{improvedbab}) of the Adam optimizer~\citep{Kingma2015}, with a learning rate of $0.1$, decayed by $0.98$ at each iteration. 
Similarly, $\alpha$-CROWN for the intermediate bounds is run with Adam for $5$ iterations, a learning rate of $1$, decayed by $0.98$ at each iteration. 
Early termination is triggered when an exponential moving average of the expected branching improvement, computed on the subproblem with the smallest lower bound within the current sub-problem batch, suggests that the decision threshold will be crossed after the timeout. The time to deplete the sub-problem queue (estimated via the runtime per bounding batch) is also added to the estimated time to termination.

\paragraph{Network architectures}
Table~\ref{tab:eth-nets} reports the details of the employed network architectures for both the training and the branching experiments.

\begin{figure}[t!]
	\begin{subfigure}{0.5\textwidth}
		\centering
		\begin{tabular}{lc}
			\textbf{Hyperparameter}&\textbf{Value}\\
			\toprule
			IBP-R with and without masking &\\
			\midrule
			Optimizer & SGD \\
			Learning rate & $10^{-2}$ \\
			Learning rate exponential decay & 0.95 \\
			Batch size & 100 \\
			Total training steps & 800 \\
			Mixing steps & 600 \\
			PGD attack steps & 8 \\
			PGD attack step size & 0.25 \\
			$\alpha$ & 2.1 \\
			$\lambda$ & $2 \times 10^{-5}$ \\
			$\frac{\rho}{2}$ & $10^{-4}$ \\
			\bottomrule
		\end{tabular}
		\caption{$\epsilon_{\text{ver}}=2/255$.}
	\end{subfigure}
	\begin{subfigure}{0.5\textwidth}
		\centering
		\begin{tabular}{lc}
			\textbf{Hyperparameter}&\textbf{Value}\\
			\toprule
			IBP-R with and without masking &\\
			\midrule
			Optimizer & SGD \\
			Learning rate & $10^{-2}$ \\
			Learning rate exponential decay & 0.95 \\
			Batch size & 150 \\
			Total training steps & 800 \\
			Mixing steps & 600 \\
			PGD attack steps & 8 \\
			PGD attack step size & 0.25 \\
			$\lambda$ & $10^{-5}$ \\
			\midrule
			IBP-R &\\
			\midrule
			$\alpha$ & 1.7 \\
			$\frac{\rho}{2}$ & $5 \times 10^{-3}$ \\
			\midrule
			IBP-R with Masking &\\
			\midrule
			$\alpha$ & 1.6 \\
			$\frac{\rho}{2}$ & $10^{-2}$ \\
			\bottomrule
		\end{tabular}
		\caption{$\epsilon_{\text{ver}}=8/255$.}
	\end{subfigure}
		\caption{\label{tab:hyperparameters} IBP-R hyper-parameters for the experiment of table~\ref{fig:cifar-tables}.}
\end{figure}

\begin{table}[b!]
	\centering \small
	\begin{adjustbox}{max width=\textwidth, center}
		\begin{tabular}{|c|c|}
			\hline
			\textbf{Network Specifications}\TBstrut & \textbf{Network Architecture} \TBstrut\\
			\hline
			\begin{tabular}{ll}
				Perturbation radius: & $\epsilon_{\text{ver}}=2/255$ \\
				Activation: & ReLU \\
				Total activations: & 49402 \\
				Total parameters: & 2133736
			\end{tabular}
			& \begin{tabular}{@{}c@{}} 
				\footnotesize Conv2d(3, 32, 3, stride=1, padding=1) \Tstrut \\
				\footnotesize Conv2d(32,32,4, stride=2, padding=1)\\
				\footnotesize Conv2d(32,128,4, stride=2, padding=1)\\
				\footnotesize linear layer of 250 hidden units \\
				\footnotesize linear layer of 10 hidden units \Bstrut
			\end{tabular}\\
			\hline
			\begin{tabular}{ll}
				Perturbation radius: & $\epsilon_{\text{ver}}=8/255$ \\
				Activation: & ReLU \\
				Total activations: & 16643 \\
				Total parameters: & 2118856  
			\end{tabular}  
			& \begin{tabular}{@{}c@{}} 
				\footnotesize Conv2d(3, 32, 5, stride=2, padding=2) \Tstrut \\
				\footnotesize Conv2d(32,128,4, stride=2, padding=1)\\
				\footnotesize linear layer of 250 hidden units \\
				\footnotesize linear layer of 10 hidden units \Bstrut
			\end{tabular}\\
			\hline
		\end{tabular}
	\end{adjustbox}
	\caption{\label{tab:eth-nets} Specifications of the employed network architectures for the experiments of $\S$\ref{sec:experiments}.}
\end{table}

\subsection{Jax Porting of COLT} \label{sec:jax-colt} 

\looseness=-1
In order to ensure a fair timing comparison with our IBP-R, we ported COLT, whose original implementation is in PyTorch~\citep{Paszke2019}, to Jax~\citep{jax2018github}.
The porting resulted in speed-up factors of around $5.7$ and $4.4$ for the $\epsilon_{\text{ver}}=2/255$ and $\epsilon_{\text{ver}}=8/255$ experiments, respectively: see table~\ref{fig:cifar-tables}. 
The standard accuracy results were similar to the original implementation (see table~\ref{fig:cifar-tables}), testifying the validity of the porting. 
The verified accuracy of our experiments is larger than the one reported in the literature: this is likely due to the different, and arguably more effective, verification procedure that we employed.

We conclude this subsection by reporting the main functional difference of our Jax porting with respect to the original implementation.
COLT's Cauchy random projections-approximated zonotope intermediate bounds require extensive use of median values computed over arrays. The median values are employed as estimators for the $\ell_1$ norm~\citep{Li2007} of the zonotope propagation matrices. Unfortunately, these operations do not scale very well in Jax~\citep{jax-median-issue}.
Therefore, we rely on a different $\ell_1$ estimator, based on the geometric mean~\citep{Li2007}, and we clip the employed Cauchy samples to ensure numerical stability during training.

\section{Supplementary Branching Experiments} \label{sec:supp-branching}

We now complement the branching results in $\S$\ref{sec:experiments-verify} by repeating the same complete verification experiment on two COLT-trained networks.

\begin{figure}[t!]
	\begin{subtable}{0.49\columnwidth}
		\sisetup{detect-weight=true,detect-inline-weight=math,detect-mode=true}
		\centering
		
		\begin{adjustbox}{max width=\textwidth, center}
			\begin{tabular}{
					l
					S[table-format=4.3]
					S[table-format=4.3]
					S[table-format=4.3]
					S[table-format=4.3]
					S[table-format=5.3]
					S[table-format=4.3]
				}
				{} & \multicolumn{3}{l}{$\epsilon_{\text{ver}}=2/255$ COLT} & \multicolumn{3}{l}{$\epsilon_{\text{ver}}=8/255$ COLT} \\
				\toprule
				\multicolumn{1}{ c }{Method} &
				\multicolumn{1}{ c }{time [s]} &
				\multicolumn{1}{ c }{subproblems$^*$} &
				\multicolumn{1}{ c }{$\%$Timeout} &
				\multicolumn{1}{ c }{time [s]} &
				\multicolumn{1}{ c }{subproblems$^*$} &
				\multicolumn{1}{ c }{$\%$Timeout} \\
				\midrule
				\textsc{UPB} &  \B  98.87 &   \B    257.17 &  \B   5.16 &  \B 100.19 &      2924.80 &     4.46 \\
				\textsc{FSB} &   104.41 &       379.77 &     5.41 &   101.00 &      \B 2840.52 &    \B 4.09 \\
				\textsc{SR} &   229.77 &      5894.53 &    12.29 &   318.21 &     20916.78 &    15.99 \\
				\bottomrule
				
				\\[-5pt]
				\multicolumn{7}{ c }{\scriptsize
					\begin{minipage}{1.5\columnwidth}
						$^*$computed on the properties that did not time out for neither UPB nor FSB. The inclusion of timed-out results in the average leads to an overestimation of the number of subproblems for the less expensive branching strategy.
					\end{minipage}
				}
				
			\end{tabular}
		\end{adjustbox}
		\vspace{-3pt}
		\subcaption{\label{table:branching-colt}\small Comparison of average runtime, average number of solved subproblems and the percentage of timed out properties. The best performing method is highlighted in bold.}
	\end{subtable}
	\begin{subfigure}{0.5\columnwidth}
		\begin{subfigure}{.49\textwidth}
			\centering
			\includegraphics[width=\textwidth]{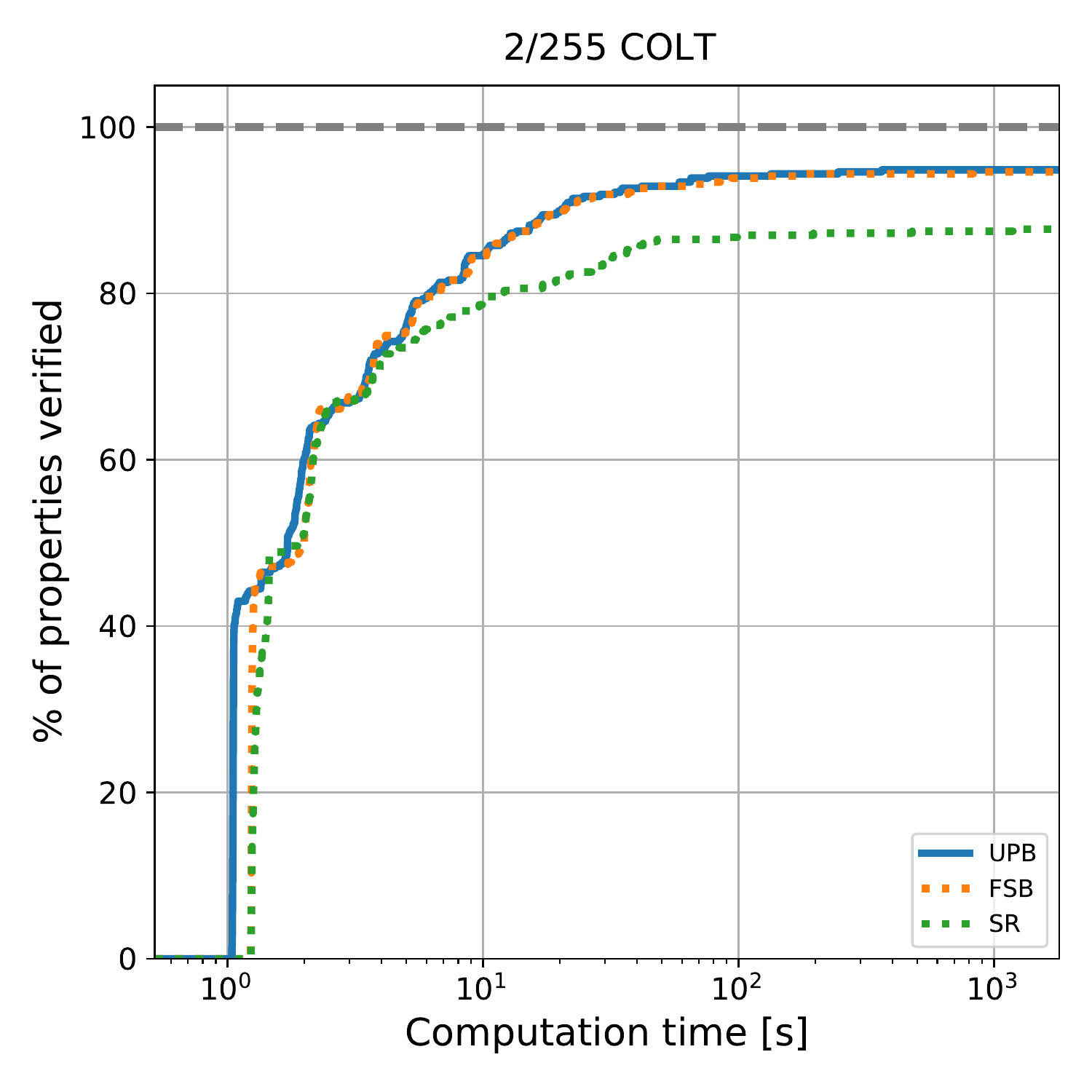}
		\end{subfigure}
		\begin{subfigure}{.49\textwidth}
			\centering
			\includegraphics[width=\textwidth]{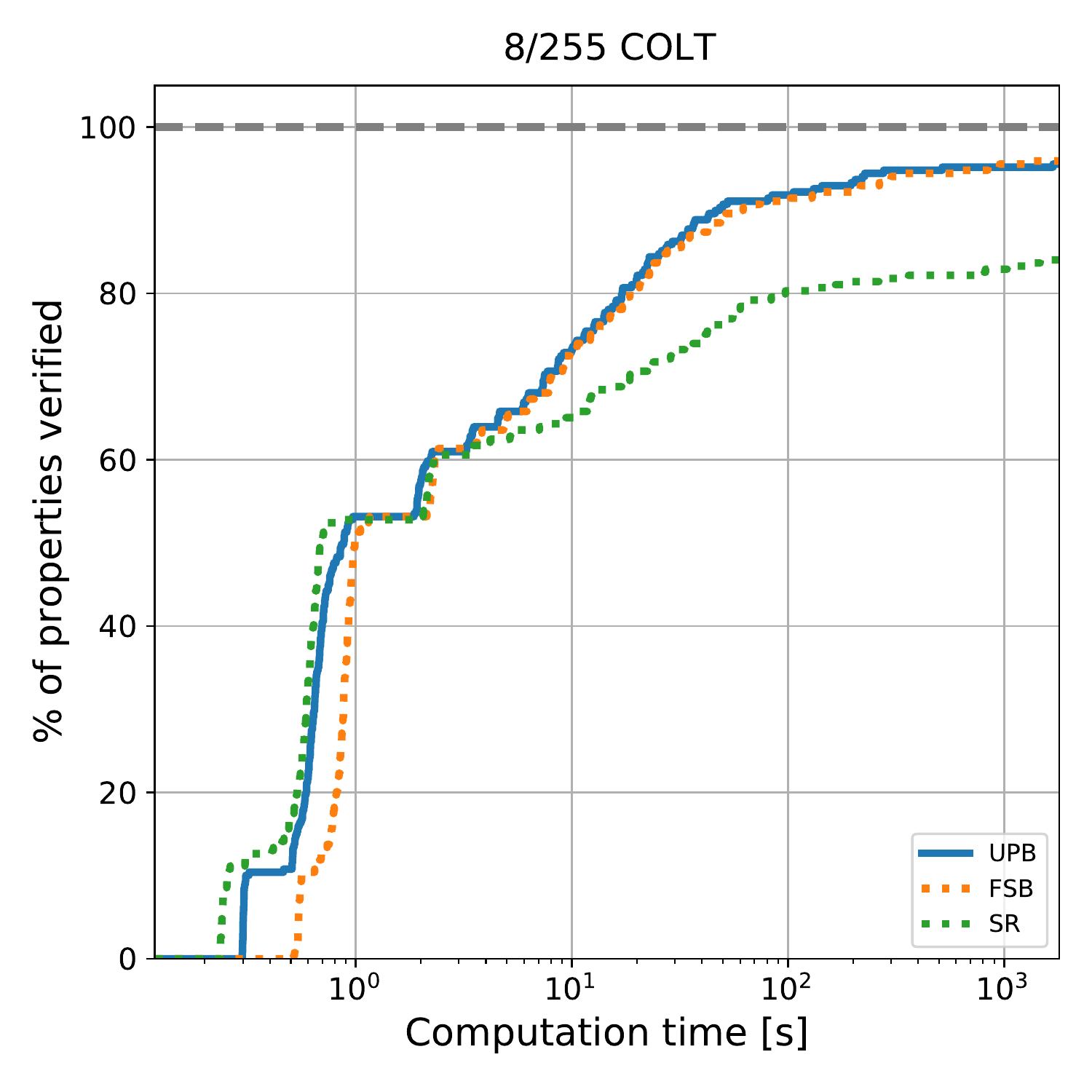}
		\end{subfigure}
		\vspace{-7pt}
		\subcaption{Cactus plots: percentage of solved properties as a function of runtime. Baselines are represented by dotted lines.}
	\end{subfigure}
	\vspace{-5pt}
	\caption{ \label{fig:branching-colt} Complete verification performance of different branching strategies, on two COLT-trained CIFAR-10 networks. }
\end{figure}

Figure~\ref{table:branching-colt} confirms the results from $\S$\ref{sec:experiments-verify}, albeit with reduced margins between UPB and FSB.
UPB yields small ($<6\%$) improvements on the average verification times with respect to FSB. 
In addition, UPB increases the average number of visited subproblems on the $\epsilon_{\text{ver}}=8/255$ network, and reduces it for $\epsilon_{\text{ver}}=2/255$, where it also decreases the number of visited subproblems and timed-out properties. 
As for the IBP-R-trained networks, UPB yields branching decisions competitive with those of FSB while incurring smaller overheads.
Finally, as seen in $\S$\ref{sec:experiments-verify} , SR is significantly slower than both UPB and FSB.

\end{document}